\pdfoutput=1

\documentclass[11pt]{article}

\usepackage{eacl2023}

\usepackage{times}
\usepackage{latexsym}

\usepackage[T1]{fontenc}

\usepackage[utf8]{inputenc}

\usepackage{microtype}

\usepackage{inconsolata}

\usepackage{graphicx}
\usepackage{amsmath,amssymb,amsfonts}
\usepackage{algorithm}
\usepackage{algorithmicx}
\usepackage{algpseudocode}
\usepackage{textcomp}
\usepackage{xcolor}
\usepackage{amsthm}
\usepackage{booktabs}
\usepackage{array}

\newtheorem{theorem}{Theorem}

\usepackage{mathtools}
\usepackage{float}
\newtheorem{assumption}{Assumption}[section]

\title{Quantifying Modality Contributions via Disentangling Multimodal Representations}

\author{Padegal Amit$^{1}$\thanks{\quad Equal Contribution}~, Omkar Mahesh Kashyap$^{1}$\footnotemark[1], Namitha Rayasam$^{1}$\footnotemark[1]\\
\textbf{Nidhi Shekhar$^{1}$, Surabhi Narayan$^{1}$}\\
$^{1}$PES University\\
 \texttt{\{padegal.amit, omkar.m.kashyap, rayasam.namitha, shekhar.nidhi\}@gmail.com},\\ 
 \texttt{\{surabhinarayan\}@pes.edu}
\\
}

\begin{document}
\maketitle
\begin{abstract}
Quantifying modality contributions in multimodal models remains a challenge, as existing approaches conflate the notion of contribution itself. Prior work relies on accuracy-based approaches, interpreting performance drops after removing a modality as indicative of its influence. However, such outcome-driven metrics fail to distinguish whether a modality is inherently informative or whether its value arises only through interaction with other modalities. This distinction is particularly important in cross-attention architectures, where modalities influence each other’s representations. In this work, we propose a framework based on Partial Information Decomposition (PID) that quantifies modality contributions by decomposing predictive information in internal embeddings into unique, redundant, and synergistic components. To enable scalable, inference-only analysis, we develop an algorithm based on the Iterative Proportional Fitting Procedure (IPFP) that computes layer and dataset-level contributions without retraining. This provides a principled, representation-level view of multimodal behavior, offering clearer and more interpretable insights than outcome-based metrics.

\end{abstract}

\section{Introduction}

\begin{figure}[ht]
  \centering
  \includegraphics[width=0.4\textwidth, height=8cm, keepaspectratio]{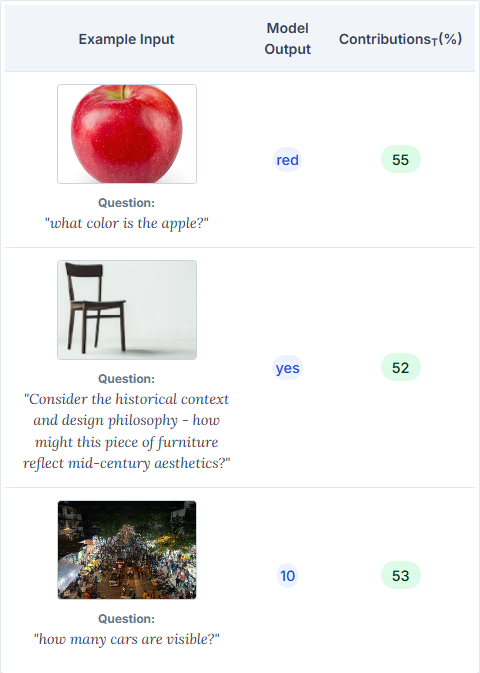}
  
  \caption{Overview of the proposed modality contribution framework, where text contribution ($C_T$) and image contribution ($C_I = 100 - C_T$) together form a normalized 100\% distribution on the BLIP model. The first row shows simple image-simple text, the second shows simple image-complex text, and the third shows complex image-simple text.}
  \label{fig:our-method-intro}
\end{figure}

Recent advances in multimodal learning have enabled models to process and align information across sensory modalities such as vision and language. However, despite impressive empirical results, current methods still struggle with true multimodal integration. They often exhibit modality imbalance, a tendency to over-rely on one modality while underutilizing the other \citet{peng2022balanced}, \citet{huang2022modality}, \citet{fan2023pmr}, 
\citet{wei2024enhancing}. As illustrated in Figure~\ref{fig:our-method-intro}, this imbalance appears asymmetrically across both vision and text.

Existing works attempt to quantify modality contribution by perturbing or masking inputs \cite{gat2021perceptual}, \cite{wenderoth2025measuring}, \cite{wang2025multishap} or by employing gradient and attention-based explanations, such as Integrated Gradients (IG) \citep{sundararajan2017axiomatic} and Grad-CAM \citep{selvaraju2017grad}. Recent works use Shapley-based approaches, such as MM-SHAP \cite{parcalabescu2022mm} and its extensions \citet{wang2025multishap}, \citet{goldshmidt2024tokenshapinterpretinglargelanguage}, \citet{Goldshmidt2025AttentionPP}. However, these methods largely treat modalities as independent sources of information, overlooking the cross-modal interactions that emerge within the model’s internal feature space. 

While the methods above often treat modalities as independent sources of information, another line of research highlights that multimodal fusion is inherently interactional. Methods such as EMAP \citet{hessel2020does}, DIME \citet{lyu2022dime}, and MultiViz \citet{liang2022multiviz} demonstrate that meaningful cross-modal dependencies play a critical role in shaping the final prediction.

To bridge these perspectives, we introduce a unified information-theoretic framework that captures both modality attribution and cross-modal interactions through PID \citep{williams2010nonnegative}. PID decomposes total predictive information into unique, redundant, and synergistic components. This principled decomposition offers clear insights into how individual modalities contribute both independently and collectively, enabling a deeper understanding of how multimodal fusion arises within the model’s internal feature representations.

Unlike previous PID-based approaches that rely on computationally intensive conic solvers or require auxiliary network training \citep{liang2023quantifying}, we present a scalable, inference-only approximation leveraging the Iterative Proportional Fitting Procedure (IPFP) \citep{bacharach1965estimating}. This makes our approach parameter-free, computationally efficient, and directly applicable for post-hoc multimodal analysis.

Our key contributions are as follows:

\begin{itemize}

    \item We propose the first modality contribution metric that jointly captures modality-specific and interactional effects.
    
    \item We develop a quantification approach based on PID and propose a novel, computationally efficient metric derived using the IPFP algorithm.
    
    \item We evaluate our method across diverse VLMs and datasets, supported by synthetic experiments and ablation studies.
\end{itemize}

\section{Related work}

\subsection{Gradient and Attention-Based Methods}
Gradient-based approaches, such as Integrated Gradients (IG) \citet{sundararajan2017axiomatic} attribute contributions through path-integrated gradients but suffer from instability, baseline sensitivity, and high attribution noise \citep{zhuo2024ig}. By relying solely on explicand gradients and neglecting counterfactual ones, IG fails to capture higher-order effects, like redundant or synergistic contributions, where features jointly influence predictions in non-additive ways, and violates key Shapley axioms, yielding spurious attributions. More broadly, gradient saliency often reflects the model’s implicit density induced by the softmax layer rather than genuine discriminative reasoning \citep{srinivas2020rethinking}.

Attention-based explanations share similar pitfalls. Attention weights may misrepresent causal polarity \citep{liu2022rethinking}, fail to correlate with true feature importance \citep{wiegreffe-pinter-2019-attention}, and can be adversarially altered without affecting predictions \citep{serrano-smith-2019-attention}. White-box visualization methods such as attention maps and Grad-CAM \citet{selvaraju2017grad} offer coarse interpretability but remain architecture-dependent and cannot disentangle synergistic from suppressive cross-modal effects. 

\subsection{Perturbation-based methods}
Early studies on quantifying modality contribution focused on uncovering dataset biases that enable strong unimodal performance \citep{goyal2017makingvvqamatter}. Subsequent work introduced perturbation-based tests to quantify modality importance by observing performance degradation when one modality is removed or altered. These approaches fall into two categories: deletion methods, which suppress or mask modality features \citep{shekhar-etal-2017-foil, madhyastha-etal-2018-defoiling, frank-etal-2021-vision}, and contradiction or foiling methods, which inject misleading inputs such as swapped captions or textual foils \citep{gat2021perceptualscoredatamodalities, shekhar-etal-2019-evaluating, parcalabescu-etal-2022-valse}. However, their reliance on accuracy-based evaluation limits interpretability, as accuracy of the models outweighs genuine cross-modal understanding. This dependence has led to inconsistent findings across studies and highlights the need for performance-agnostic approaches that more reliably capture true modality reliance.

\subsection{Shapley-based methods}

Shapley-based methods, grounded in Shapley theory \cite{shapley1953value}, provide a principled framework for attributing model predictions by averaging each feature’s marginal contribution over all possible subsets of features \cite{lundberg2017unified}.
TokenSHAP \citet{goldshmidt2024tokenshapinterpretinglargelanguage} and PixelSHAP \citet{Goldshmidt2025AttentionPP} extend this framework to tokens and image regions, respectively, yielding fine-grained unimodal attributions but missing cross-modal dependencies. MultiSHAP \cite{wang2025multishap} models patch-token interactions to separate synergistic and suppressive effects, but it incurs high Monte Carlo cost. MM-SHAP \citep{parcalabescu2022mm} introduces a performance-agnostic metric that aggregates absolute Shapley contributions per modality, however it captures only aggregate modality proportions, discarding directional and interaction information and remaining sensitive to masking artifacts. 

\begin{figure*}[ht]
  \centering
  \includegraphics[width=\textwidth]{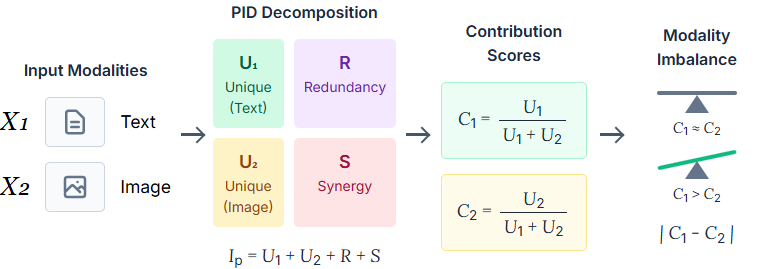} 
  \caption{Overview of the proposed modality contribution framework, where text contribution ($C_T$) and image contribution ($C_I = 100 - C_T$) together form a normalized 100\% distribution.}
  \label{fig:our-method}
\end{figure*}

\section{Methodology}

PID extends classical information theory to multiple sources by decomposing the total mutual information \( I_p(X_1, X_2; Y) \), where \(X_1, X_2\), and \(Y\) represent the distributions of two sources and the target variable, respectively, into four distinct components:

\begin{itemize}
\item \textbf{Redundant Information ($R$)}: Information about \(Y\) that is shared by both sources \(X_1\) and \(X_2\):
\begin{equation}
\begin{aligned}
R &= \max_{q \in \Delta_p} I_q(X_1; X_2; Y)
\end{aligned}
\label{eq:defn_r}
\end{equation}
\item \textbf{Unique Information from Source 1 ($U_1$)}: The information about \(Y\)  provided exclusively by source \(X_1\), without overlap from source \(X_2\):
\begin{equation}
\begin{aligned}
U_1 &= \min_{q \in \Delta_p} I_q(X_1; Y \mid X_2)
\end{aligned}
\label{eq:defn_u1}
\end{equation}
\item \textbf{Unique Information from Source 2 ($U_2$)}: The information about \(Y\) provided exclusively by source \(X_2\), without overlap from source \(X_1\):
\begin{equation}
\label{eq:defn_u2}
\begin{aligned}
U_2 &= \min_{q \in \Delta_p} I_q(X_2; Y \mid X_1),
\end{aligned}
\end{equation}
\item \textbf{Synergistic Information ($S$)}: The information that arises only when both sources \(X_1\) and \(X_2\) are considered jointly, capturing their interaction:
\end{itemize}
\begin{equation}
\begin{aligned}
S &= I_p(X_1, X_2; Y) - \min_{q \in \Delta_p} I_q(X_1, X_2; Y)
\end{aligned}
\label{eq:defn_s}
\end{equation}

These four components satisfy the fundamental decomposition: \[I_p(X_1, X_2; Y) = R + U_1 + U_2 + S
\]
This decomposition explicitly disentangles individual modality contributions 
($U_1$, $U_2$) from cross-modal interactions ($R$, $S$). Together, these components provide a principled way to disentangle modality-specific and joint information contributions within multimodal fusion.

\subsection{Computing PID}

Let $X_1 \in \mathcal{X}_1$ and $X_2 \in \mathcal{X}_2$ denote the text and image embeddings, respectively, and $Y \in \mathcal{Y}$ represent the output embeddings. Define the simplex of all joint distributions,

\[
\Delta = \mathcal{P}(X_1 \times X_2 \times Y)
\]
and the affine feasible set.

\begin{equation}
\begin{aligned}
\Delta_p = \big\{ q \in \Delta :\ 
& q(x_i, y) = p(x_i, y), \\
& \forall x_i \in \mathcal{X}_i, \; y \in \mathcal{Y},\ i \in \{1,2\}\big\}
\end{aligned}
\label{eq:delta_p}
\end{equation}
which preserves the observed modality-target marginals $p(x_1, y)$ and $p(x_2, y)$.

Following \citet{liang2023quantifying}, the PID components are obtained via a KL-projection formulation:

\begin{equation}
\label{eq:conditional_entropy}
\begin{aligned}
H_q(Y \mid X_1, X_2) &= \log |\mathcal{Y}| - \mathrm{KL}\big(q \,\|\, \tilde{q}\big), \\
\text{where} \quad \tilde{q}(x_1, x_2, y) &= \frac{q(x_1, x_2)}{|\mathcal{Y}|}
\end{aligned}
\end{equation}
 Here, the joint distribution \( q(x_1, x_2, y) \) can be represented as a tensor \( Q \) of shape \( |X_1| \times |X_2| \times |\mathcal{Y}| \).
 However, optimizing over the full joint tensor
\(
Q 
\)
is computationally expensive. We discuss this in more detail in Appendix~\ref{sec:appendixE}

To address this, we first consider the corresponding convex problem that solves Equation~\eqref{eq:conditional_entropy}.

\begin{equation}
\begin{aligned}
& \arg \min_{Q, \tilde{Q}} F(Q, \tilde{Q}) = \mathrm{KL}(Q \,\Big\|\, \tilde{Q}) \\
\text{s.t.} \quad & \tilde{Q}(x_1, x_2, y) = \frac{Q(x_1, x_2)}{|\mathcal{Y}|} \quad \forall x_1, x_2, y, \\
& \sum_{x_2} Q(x_1, x_2, y) = p(x_1, y), \quad \forall x_1, y, \\
& \sum_{x_1} Q(x_1, x_2, y) = p(x_2, y), \quad \forall x_2, y, \\
& Q \geq 0, \quad \forall x_1, x_2, y, \\
& \sum_{x_1, x_2, y} Q(x_1, x_2, y) = 1.
\end{aligned}
\label{eq:opt_problem}
\end{equation}

The objective \( F(Q, \tilde Q) \) is jointly convex in both arguments, and all constraints are affine.
Hence, Equation~\eqref{eq:opt_problem} defines a well-posed convex program over the positive orthant.

\begin{theorem}[Existence and Uniqueness of \( Q^\star \)]
\label{thm:existence-uniqueness}
The subproblem in Equation ~\ref{eq:opt_problem}
\[
Q^\star = \arg\min_{Q \in \Delta_p} \mathrm{KL}(Q \,\|\, \tilde Q)
\]
admits a unique minimizer \( Q^\star \in \Delta_p \).
\end{theorem}

\begin{proof}
Under the assumptions outlined in Appendix ~\ref{sec:appendixA}, the set \( \Delta_p \) is compact and convex, which guarantees that the KL projection is both well-defined and unique. This justifies the use of the I-projection results from \citet{csiszar2004information}, which we leverage to prove the theorems. These results are further detailed in Appendix~\ref{thm:existence-uniqueness}.
\end{proof}
Thus, Theorem~\ref{thm:existence-uniqueness} establishes that our formulation yields the same optimizer as that of \citet{liang2023quantifying}, ensuring consistency with prior work.

\subsection{Alternating Minimization via IPFP}
\label{sec:alternating-min}
We solve Equation~\eqref{eq:opt_problem} by alternating KL-projections,
following the structure of IPFP. At each iteration \( t \), we update \( Q \) and \( \tilde Q \) as follows.

\paragraph{(1) Q-step (I-projection).}

Given the current iterate \( \tilde Q^{(t)} \), update:
\[
Q^{(t+1)} = 
\arg\min_{Q \in \Delta_p}
\mathrm{KL}(Q \,\|\, \tilde Q^{(t)}).
\]
This projection enforces the marginal constraints
\( p(x_1, y) \) and \( p(x_2, y) \) while staying closest to \( \tilde Q^{(t)} \)
under the KL divergence.
Since the constraints in \( \Delta_p \) factorize over labels \( y \),
the problem decomposes into independent subproblems
\[
Q_y^{(t+1)} = \arg\min_{Q_y \in \Delta_{p(y)}} 
\mathrm{KL}(Q_y \,\|\, \tilde Q_y^{(t)}),
\]
each corresponding to a label-specific KL-regularized optimal transport problem.
Each of these can be efficiently solved using
Sinkhorn iterations \citep{cuturi2013sinkhorn}.

\paragraph{(2) \(\tilde Q\)-step (closed-form update).}
Given the updated coupling \( Q^{(t+1)} \), the auxiliary variable is updated as
\[
\tilde Q^{(t+1)}(x_1, x_2, y)
= \frac{Q^{(t+1)}(x_1, x_2)}{|\mathcal{Y}|},
\]
which is the closed-form solution minimizing
\( \mathrm{KL}(Q^{(t+1)} \,\|\, \tilde Q) \)
under the normalization constraint. This alternation monotonically decreases \( F(Q, \tilde{Q}) \) and converges globally.

\begin{theorem}[Global convergence of alternating KL-projections]
\label{thm:convergence}
Let \( (Q^{(t)}, \tilde Q^{(t)}) \) be the iterates produced by the alternating minimization (See Section ~\ref{eq:opt_problem}).
Define \( F(Q, \tilde{Q}) \coloneqq \mathrm{KL}(Q \,\|\, \tilde{Q}) \). Then
\[
F(Q^{(t+1)}, \tilde Q^{(t+1)})
\le F(Q^{(t)}, \tilde Q^{(t)}) \quad \forall t,
\]
and the sequence \( \{F(Q^{(t)}, \tilde Q^{(t)})\} \)
converges monotonically to a unique global minimum \( F^\star \ge 0 \).
Moreover, \( Q^{(t)} \to Q^\star \), where \(Q^\star\) is the unique minimizer of Equation~\eqref{eq:opt_problem} and \(
\tilde{Q}(t) \to \tilde{Q}^\star \), the corresponding projection of \( \tilde Q\) onto 
\(Q_y \in \Delta_p(y)\)
\end{theorem}
\begin{proof}
The result follows from the alternating divergence minimization theorem
of \cite{csiszar2004information},
applied to the two convex sets defined by the
Q-step and \(\tilde Q\)-step projections.
Monotonic decrease follows from the three-point property of KL divergence,
and convergence from compactness and strict convexity of the KL divergence. A detailed proof is provided in Appendix~\ref{sec:appendixB}.
\end{proof}

Thus, Theorem~\ref{thm:convergence} guarantees that the alternating I-projection procedure
is globally convergent and initialization-independent,
with each iteration strictly decreasing the objective until reaching the fixed point \( Q^\star \).

Direct IPFP (Sinkhorn) updates may suffer from numerical instability due to repeated exponentiation and multiplication of small values.
Following \citet{peyre2019computational},
we perform all updates in the log domain for stability. 
For each label \( y \in Y
\), define the positive kernel \( A_y(x_1,x_2) = \tilde Q(x_1,x_2,y) > 0 \)
and marginals \( r_y(x_1) = p(x_1,y) \), \( c_y(x_2) = p(x_2,y) \).

The stabilized Sinkhorn updates are:
\begin{equation}
\begin{aligned}
\log u_y &\leftarrow \log r_y \;-\; \operatorname{LSE}_{x_2}\!\big(\log A_y + \log v_y\big),\\[2pt]
\log v_y &\leftarrow \log c_y \;-\; \operatorname{LSE}_{x_1}\!\big(\log A_y + \log u_y\big),
\end{aligned}
\label{eq:log_updates}
\end{equation}
where \( \mathrm{LSE}(\cdot) \) denotes the log-sum-exp operator,
ensuring numerical stability.
After convergence,
\begin{equation}
\begin{split}
R_y &= \exp\!\big(\log A_y + \log u_y + \log v_y\big) \\
    &= \operatorname{diag}(u_y)\, A_y\, \operatorname{diag}(v_y),
\end{split}
\label{eq:Ry_factorization}
\end{equation}
which satisfies the marginal constraints exactly. We also include the Algorithm of our IPFP approach in Appendix~\ref{sec:appendixC}.

\subsection{Modality Contribution Metric}
Having obtained scalable estimates of the PID components via our IPFP-based solver, we now construct interpretable modality contribution metrics that directly quantify how much each modality contributes to model predictions.

We define the normalized contribution of modality $i$ as
\begin{equation}
    C_i = \frac{U_i}{\sum_j U_j},
\label{eq:Ci_normalized}
\end{equation}
where $U_i$ is the unique information provided by modality $i$. The sum in the denominator normalizes the contributions, ensuring they are expressed as fractions of the total unique information across all modalities. This normalization allows $C_i$ to represent the relative importance of each modality, independent of the absolute scale of unique information, and ensures that the contributions sum to 1.
We explain the complete framework in Figure~\ref{fig:our-method}.

\subsection{Why PID enables a contribution score}
By isolating the unique information ($U_1, U_2$) from both shared and synergistic effects, PID provides a clear distinction between information contributed independently by each modality and that which emerges from cross-modal interactions. This separation helps prevent conflating individual and interactional contributions. Importantly, PID is accuracy-agnostic, meaning it quantifies information content regardless of whether the model’s predictions are correct \citep{parcalabescu2022mm}. Recent work has also emphasized the limitations of perturbation-based methods, advocating for approaches that analyze a model's internal representations rather than relying solely on its outputs \citep{sim2025can}. PID naturally aligns with this perspective, as it operates directly on the model’s internal feature distributions. This allows for the computation of modality contribution scores while capturing cross-modal dynamics, without requiring model perturbations.

\section{Implementation Details}

Following \citet{liang2023quantifying}, we estimate PID by first converting the continuous modality embeddings into discrete distributions with finite support. To do this, we apply feature binning and clustering techniques. Specifically, for high-dimensional embeddings, we use K-means clustering to obtain discrete support sets \(|X_1|, |X_2|\) and \(Y\) for the two modalities and the target.

The IPFP solver runs for up to 50 iterations. Within each iteration, it performs a maximum of 100 Sinkhorn updates in the log domain (see Equation \ref{eq:log_updates}) per label. The iterative process halts early if either the marginal deviation or the relative change in the KL objective
\(
F(Q, \tilde{Q})
\) drops below a threshold of \(1 \times 10^{-8}\). 

\section{Experiments}

\subsection{Synthetic dataset}
Our first goal is to evaluate the accuracy of our proposed estimators with respect to the ground truth or human judgment, as done by \citet{liang2023quantifying}.
We sample from a binary bitwise distribution where each bitwise operator's PID can be solved exactly when $y$ and \( x_i \)'s are discrete and low-dimensional.

\begin{figure}[ht]
    \centering
    \includegraphics[width=0.48
    \textwidth]{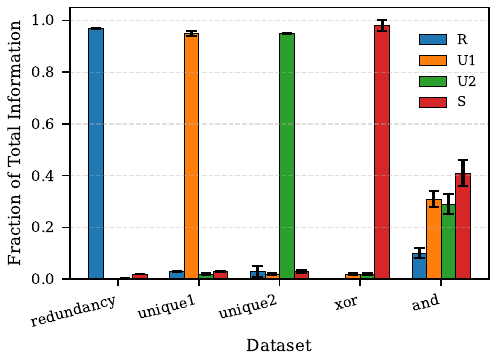} 
    \caption{Comparison of proposed estimators with true PID values on synthetic datasets based on binary bitwise operations. R = Redundancy, U1 \& U2 = Unique information, S = Synergy.}
    \label{fig:sample-ground-truth}
\end{figure}

The \textit{and} dataset (\( y = x_1 \wedge x_2 \)) exhibits moderate synergy, as the output is 1 only when both inputs are 1, requiring joint information. It has little redundancy since neither input alone reliably predicts \( y = 1 \), but some unique information is present because if either input is 0, the output is 0. 
 
In the \textit{xor} dataset (\( y = x_1 \oplus x_2 \)), synergy is high because neither input alone predicts \( y \), and there is no redundancy or unique information in the individual inputs; only their combination reveals the output. 
The \textit{unique1} dataset (\( y = x_1 \)) contains no synergy or redundancy because \( x_2 \) adds no information, and all information about \( y \) is unique to \( x_1 \). The \textit{unique2} dataset (\( y = x_2 \)) mirrors \textit{unique1}, but with the roles of the inputs reversed: all information about \( y \) is uniquely provided by \( x_2 \), while \( x_1 \) contributes nothing. As such, it contains no redundancy or synergy.
Lastly, the \textit{redundancy} dataset, where both inputs are identical (\( x_1, x_1 \)) and \( y = x_1 \), has no synergy but maximum redundancy since both predictors share the same information about \( y \). Figure \ref{fig:sample-ground-truth} illustrates that our estimated PID components are consistent with the theoretical characteristics of these synthetic datasets, providing evidence of the estimator’s validity.

\subsection{Synthetic Validation of Fusion Mechanisms}
\begin{figure}[ht]
  \centering
  \includegraphics[width=0.95\linewidth]{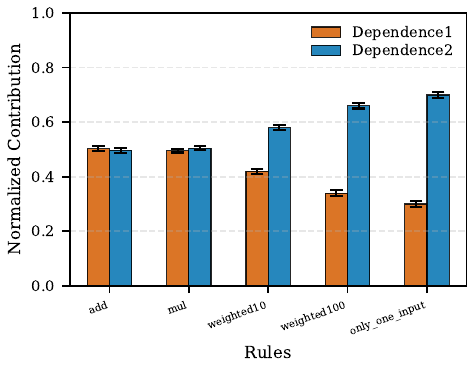}  
  \caption{Synthetic validation of the proposed PID-based metric based on rules: additive, multiplicative, weighted additive and single input.}
  \label{fig:synthetic-gaussian}
\end{figure}

To verify that the proposed metric behaves as expected under controlled conditions, we conduct a synthetic validation experiment using two independent Gaussian variables: 
\( X_1 \sim \mathcal{N}(0, 1) \) and \( X_2 \sim \mathcal{N}(0, 1) \)
To ensure statistical independence, the inputs are decorrelated using Independent Component Analysis (ICA) 
\cite{hyvarinen2001independent}. The synthetic output $Y$ is generated under several predefined fusion rules: \textit{add}($Y = X_1 + X_2$), \textit{mul} ($Y = X_1 \times X_2$), \textit{weighted 10} ($Y = X_1 + 10X_2$), \textit{weighted 100}($Y = X_1 + 100X_2$) and
\textit{only one input}($Y = X_2$). These formulations emulate different multimodal interactions, from balanced contributions to strongly biased or multiplicative relationships.
Because the ground-truth modality contributions are analytically known, this setup provides a reliable way to validate the metric’s correctness. As expected, additive and multiplicative fusion produce approximately equal contributions from both modalities (around 50\%), while weighted fusion shifts the contributions toward the modality with the larger coefficient, as shown in Figure~\ref{fig:synthetic-gaussian}. These results confirm that the proposed metric reliably captures true modality influence and accurately reflects the underlying fusion dynamics rather than random variation.

\begin{table*}[t]
\centering
\resizebox{\textwidth}{!}{
\begin{tabular}{lcc|cc|cc|cc|cc|cc}
\toprule
\textbf{Model} 
& \multicolumn{2}{c|}{\textbf{VQA}} 
& \multicolumn{2}{c|}{\textbf{GQA}} 
& \multicolumn{4}{c|}{\textbf{COCO}} 
& \multicolumn{2}{c|}{\textbf{ScienceQA}} 
& \multicolumn{2}{c}{\textbf{NLVR2}} \\
\cmidrule(lr){2-3}
\cmidrule(lr){4-5}
\cmidrule(lr){6-9}
\cmidrule(lr){10-11}
\cmidrule(lr){12-13}
 & Text & Image 
 & Text & Image 
 & \multicolumn{2}{c|}{Original} 
 & \multicolumn{2}{c|}{FOIL-COCO} 
 & Text & Image 
 & Text & Image \\
\cmidrule(lr){6-7}
\cmidrule(lr){8-9}
 & & & & & Text & Image & Text & Image & & & & \\
\midrule
PaliGemma2-3B & 53.60 & 46.39 & 56.44 & 43.55 & 55.10 & 44.90 & 54.80 & 45.20 & 53.29 & 46.70 & 48.63 & 51.36 \\
SmolVLM-500M   & 60.90 & 39.10 & 58.30 & 41.60 & 59.80 & 40.10 & 60.10 & 39.80 & 53.30 & 46.60 & 65.18 & 34.81 \\
LLaVA-1.5-7B     & 59.16 & 40.84 & 57.90 & 42.10 & 58.50 & 41.50 & 58.70 & 41.30 & 54.20 & 45.80 & 52.40 & 47.60 \\
BLIP     & 55.45 & 44.55 & 54.58 & 45.42 & 58.94 & 41.05 & 57.32 & 42.68 & 52.90 & 47.10 & 53.80 & 46.10 \\
\bottomrule
\end{tabular}
}
\caption{Modality-specific score (\%) of various models across multiple benchmarks. The COCO dataset is subdivided into Original and FOIL-COCO settings \cite{shekhar2017foil}, each reporting results with text-only and image-only input.}
\label{tab:modality-ablation}
\end{table*}

\subsubsection{Experimental Setup}

We apply our method on pretrained models without any further training. For each model, we extract the relevant intermediate representations and compute the PID decomposition, yielding per-sample contributions from each modality. This approach enables comparison of modality contributions across different tasks, ensuring the results reflect true model behavior rather than dataset-specific biases or random fluctuations \cite{parcalabescu2022mm}.

\subsubsection{Models}

We evaluate our metric across models that differ in both fusion strategy and model scale. Specifically, we consider two primary fusion paradigms: (1) cross-attention-based fusion, which enables dynamic interaction between modalities, and (2) concatenation-based fusion, which combines modality embeddings in a shared feature space without explicit cross-modal exchange. BLIP \citep{li2022blip}, a relatively compact 3B-parameter model, exemplifies the cross-attention approach through its Query Transformer (Q-Former), which bridges frozen image and text encoders via lightweight query-based attention for fine-grained alignment. In contrast, LLaVA-1.5 \citep{liu2023visual}, a larger 7-B parameter model, and PaliGemma2 \citep{steiner2024paligemma}, also with 3B parameters, employ concatenation-based fusion within their architectures. SmolVLM
\citep{marafioti2025smolvlm}, with 500M parameters, serves as an efficiency-oriented variant of the same paradigm. This range, from different fusion paradigms to different model sizes, allows us to examine how fusion mechanisms and model capacity jointly influence modality contributions and cross-modal interactions. All models were evaluated using 4-bit quantization to reduce memory footprint and computational cost while maintaining performance.

\subsubsection{Tasks and Datasets}

We evaluate our metric on five multimodal benchmarks covering diverse modality dependencies. VQA \citep{goyal2017making} and GQA \citep{hudson2019gqa} represent balanced multimodal tasks that require joint visual-textual reasoning, making them ideal for testing whether our metric captures equal modality influence. NLVR2 \citep{suhr2018corpus} focuses on visual reasoning, where models must determine whether textual descriptions align with paired images, providing a setting to evaluate cross-modal grounding and potential visual dominance. ScienceQA \citep{saikh2022scienceqa} features variable modality reliance, as different question types depend more heavily on either text or visuals, allowing us to assess sensitivity to task-specific modality shifts. Finally, FOIL-COCO \citep{shekhar2017foil} introduces controlled textual discrepancies, testing whether the metric can detect how subtle linguistic perturbations alter modality contributions.

\subsubsection{Results}
Across VQA and GQA, all major models like BLIP, LLaVA, and PaliGemma show consistent modality attributions, with text contributing around 55\% and image 45\%, indicating a balanced integration of linguistic and visual information (Table ~\ref{tab:modality-ablation}). This alignment across architectures confirms that our metric captures a stable and interpretable modality balance, consistent with prior findings that both datasets encourage joint reasoning over text and image content. SmolVLM, however, exhibits a slightly higher text bias (60\%), likely reflecting its smaller capacity and lighter visual backbone. The near-equal contributions observed for larger models can also be attributed to their VQA-style training objectives, which emphasize textual regularities while maintaining visual grounding \citep{goyal2017makingvvqamatter}. Overall, these results validate that balanced benchmarks like VQA and GQA yield moderate, near-equal modality contributions when analyzed through our information-theoretic framework.

On ScienceQA, modality contributions show greater variability, with text slightly dominating (\(\approx 53\text{-}55\%\)) due to the dataset’s heterogeneous question types. BLIP and SmolVLM exhibit more balanced behavior, indicating better adaptation to visual reasoning, whereas LLaVA and PaliGemma remain more text-oriented.

For NLVR2, a visually grounded reasoning task, most models exhibit an increase in image contribution, consistent with its emphasis on spatial and relational understanding. However, SmolVLM remains notably text-oriented (\(\approx 65\%\) text vs.\ 35\% image), suggesting limited visual grounding likely due to its smaller capacity and lightweight visual encoder. In contrast, BLIP, LLaVA, and PaliGemma demonstrate a more balanced distribution between modalities, aligning with their larger architectures and more expressive fusion mechanisms.

On the FOIL-COCO dataset, we note that the modality contribution scores do not deviate significantly from the baseline, indicating that minor textual perturbations do not drastically alter the model’s overall modality balance. This behavior is consistent with observations from \cite{parcalabescu2022mm}, which similarly reported stability under subtle input modifications. Among models, PaliGemma continues to show relatively lower text contribution, while SmolVLM exhibits a stronger bias toward text, reinforcing trends observed in other benchmarks.

\section{Ablation}
\subsection{Fusion Method Ablation}
We evaluated the impact of different multimodal fusion strategies on modality contributions. Table \ref{tab:fusion_ablation} presents the average contributions of image and text for four fusion methods: \textit{Img$\rightarrow$Text} (cross-attention from image to text, where the image guides the processing of text), \textit{Txt$\rightarrow$Image} (cross-attention from text to image, where the text guides the processing of the image), \textit{Concatenation} (which combines image and text features into a single representation vector without direct interaction), and \textit{Bidirectional Fusion} (where the model simultaneously attends to both modalities, allowing mutual influence).

\begin{table}[h]
\centering
\begin{tabular}{lcc}
\toprule
Fusion & Image & Text\\
\midrule
Img$\rightarrow$Txt & 55.3 & 44.7 \\
Txt$\rightarrow$Img & 38.5 & 61.5 \\
Concatenation               & 45.6 & 54.4 \\
Bidirectional Fusion          & 54.7 & 45.3 \\
\bottomrule
\end{tabular}
\caption{Ablation study comparing different multimodal fusion methods and their impact on modality contributions.}
\label{tab:fusion_ablation}
\end{table}

From the results, we observe that the direction of cross-attention significantly affects the balance of modality contributions. Specifically,  \textit{Img$\rightarrow$Text} slightly favors the image (55.3\%), while \textit{Txt$\rightarrow$Img}
overemphasizes the text (61.5\%), potentially underestimating image's contribution. In contrast, \textit{Concatenation} and \textit{Bidirectional Fusion} achieve a more balanced distribution, indicating a stable and equitable utilization of both modalities. These findings suggest that symmetric 
fusion methods, like concatenation and bi-directional fusion, may be preferable in scenarios where both modalities are expected to contribute 
equally, and that text to image cross-attention may introduce systematic bias toward text.

\subsection{Modality perturbations}
To further investigate the sensitivity of the model to modality-specific perturbations, we conducted an ablation by varying the textual and image input. We examined two forms of text corruption: (1) randomly shuffling word order and (2) replacing text with dummy content (\texttt{Lorem Ipsum}). For the image input, we applied Gaussian noise to it.  Table~\ref{tab:perturb_ablation} 
summarizes the mean image and text contributions for each condition.

\begin{table}[h]
\centering
\begin{tabular}{lcc}
\toprule
Perturbation & Image & Text\\
\midrule
Original      & 42.6 & 57.4 \\
Noisy image & 44.9 & 55.1 \\
Dummy text    & 47.7 & 52.3 \\
Jumbled text  & 38.8 & 61.2 \\
\bottomrule
\end{tabular}
\caption{
Ablation study showing the mean of image and text contributions under different input perturbations.
}
\label{tab:perturb_ablation}
\end{table}

The results show that perturbations consistently shift modality contributions. Adding noise to the images slightly increases image dependence, indicating the model adapts to the altered input. Replacing the text with dummy content or jumbling the word order increases text reliance, reflecting the model's sensitivity to disruptions in the linguistic signal. Notably, the increase in text contribution for jumbled inputs underscores the role of positional encoding in maintaining semantic structure.

\section{Conclusion}
We present a principled framework for quantifying modality contributions in multimodal models using PID. Our method operates directly on internal representations, enabling the separation of unique, redundant, and synergistic information, and distinguishing inherent modality importance from cross-modal interactions. Through validation on synthetic data, diverse benchmarks, and models such as BLIP, LLaVA, PaliGemma, and SmolVLM, we demonstrate that our metric produces consistent, interpretable patterns that align with known dataset biases. These results establish our approach as a scalable, model-agnostic, and performance-independent tool for analyzing multimodal interactions.

\section{Limitations}
While our PID-based framework provides a principled lens for analyzing modality contributions, it is not without limitations. First, the PID estimates are sensitive to noise in the embedding space, as small variations in feature distributions can influence the inferred information components. Second, as with other recent analyses, the method may not be fully immune to spurious correlations in the data or model representations, which could affect the attribution scores. Finally, the approach assumes access to intermediate model embeddings, which may not always be available for closed-source or API-restricted systems. Addressing these challenges, particularly improving robustness to noise and mitigating the influence of spurious correlations, remains an important direction for future work.

\bibliography{anthology, custom}
\bibliographystyle{acl_natbib}

\clearpage
\onecolumn

\appendix

\section*{Appendix A \quad Theoretical Foundations of the IPFP Framework}
\label{sec:appendixA}

\subsection*{A.1 \quad Verification of Assumptions}
\label{assumption-verification}

Let \(\mathcal X_1\), \(\mathcal X_2\), \(\mathcal Y\) be finite sets, and let $p(x_1, y)$ and $p(x_2, y)$ denote the prescribed marginals. To ensure that the alternating KL-projection updates in the IPFP procedure are well-defined and numerically stable, we assume the following standard conditions.

\begin{assumption}[Feasibility]
\label{assump:A1}
The marginal constraints are jointly realizable:
\[
\Delta_p := \left\{ 
    Q \in \mathbb{R}_{\ge 0}^{|\mathcal{X}_1| \times \mathcal{X}_2| \times |\mathcal{Y}|}
    :
    \sum_{x_2} Q(x_1, x_2, y) = p(x_1, y), \;
    \sum_{x_1} Q(x_1, x_2, y) = p(x_2, y)
\right\} \ne \emptyset.
\]
This ensures that a joint distribution satisfying all marginal constraints exists, that is, the projection problem is well-defined.
\end{assumption}

\begin{assumption}[Strict Positivity of the reference kernel.]
\label{assump:A2}
\begin{equation}
        \tilde{Q}(x_1, x_2, y) > 0 
        \quad 
        \forall (x_1, x_2, y) \in X_1 \times X_2 \times Y.
        \label{eq:positivity}
    \end{equation}
This avoids divisions by zero in multiplicative updates and guarantees that the logarithm in the KL divergence is well-defined.
\end{assumption}
\begin{assumption}[Full label support.]
\label{assump:A3}
    \begin{equation}
        p(y) > 0 
        \quad 
        \forall y \in \mathcal{Y}.
        \label{eq:labelmass}
    \end{equation}
   This prevents degenerate label slices \(Q(\cdot, \cdot |y)\) from being undefined and ensures each label participates in optimization.
\end{assumption}

Since \(\mathcal X_1\), \(\mathcal X_2\), \(\mathcal Y\) are finite, the probability simplex is compact, and $\Delta_p$ is nonempty, closed, and convex under Assumption~\ref{assump:A1}. Together with Assumption~\ref{assump:A2}, which ensures strict positivity of $\tilde{Q}$, the KL projection problem in Equation~\eqref{eq:opt_problem} is well-defined.

\section*{Appendix B \quad Proofs of Theorems}
\label{sec:appendixB}   

\begin{theorem}[Existence and Uniqueness of \( Q^\star \)]
\label{thm:existence-uniqueness}
The subproblem in Equation~\eqref{eq:opt_problem}
\[
Q^\star = \arg\min_{Q \in \Delta_p} \mathrm{KL}(Q \,\|\, \tilde Q)
\]
admits a unique minimizer \( Q^\star \in \Delta_p \).
\end{theorem}

\begin{proof} 

Since \(\mathcal X_1\), \(\mathcal X_2\), \(\mathcal Y\) are finite, the probability simplex is compact. Under Assumption~\ref{assump:A1}, the feasible set $\Delta_p$ is nonempty, closed, and convex. By Assumption~\ref{assump:A2}, $\tilde{Q} > 0$, and therefore the map
\(
Q \mapsto KL(Q \,\|\, \tilde{Q})
\)
is strictly convex in $Q$ \cite{cover1999elements} and lower semicontinuous. By the Weierstrass Extreme Value Theorem, a minimizer exists, and strict convexity implies that it is unique.

The alternating updates of IPFP correspond to successive Bregman projections onto the convex sets enforcing the marginal constraints. By the Csisz\'ar-Tusn\'ady alternating minimization theorem \cite{csiszar1984information}, the iterates $Q^{(t)}$ converge to the unique minimizer $Q^\star$, and the sequence of objective values satisfies
\[
KL\!\left(Q^{(t)} \,\|\, \tilde{Q}\right) \;\downarrow\; KL\!\left(Q^\star \,\|\, \tilde{Q}\right).
\]
That is, the KL objective is monotone nonincreasing along the iterates and converges to its optimal value.
\end{proof}

\begin{theorem}[Global convergence of alternating KL-projections]
\label{thm:convergence-proof}
Let \( (Q^{(t)}, \tilde Q^{(t)}) \) be the iterates produced by the alternating minimization (See Section ~\ref{sec:alternating-min}).
Define \( F(Q, \tilde{Q}) \coloneqq \mathrm{KL}(Q \,\|\, \tilde{Q}) \). Then
\[
F(Q^{(t+1)}, \tilde Q^{(t+1)})
\le F(Q^{(t)}, \tilde Q^{(t)}) \quad \forall t,
\]
and the sequence \( \{F(Q^{(t)}, \tilde Q^{(t)})\} \)
converges monotonically to a unique global minimum \( F^\star \ge 0 \).
Moreover, \( Q^{(t)} \to Q^\star \), where \(Q^\star\) is the unique minimizer of Equation~\eqref{eq:opt_problem} and \(
\tilde{Q}(t) \to \tilde{Q}^\star \), the corresponding projection of \( \tilde Q\) onto 
\(Q_y \in \Delta_p(y)\)
\end{theorem}

\begin{proof}
Each alternating step of IPFP performs an exact Bregman (KL) projection onto a closed convex constraint set. Therefore,
\[
F\!\left(Q^{(t+1)}, \tilde{Q}^{(t)}\right)
  \;\le\;
F\!\left(Q^{(t)}, \tilde{Q}^{(t)}\right),
\qquad
F\!\left(Q^{(t+1)}, \tilde{Q}^{(t+1)}\right)
  \;\le\;
F\!\left(Q^{(t+1)}, \tilde{Q}^{(t)}\right),
\]
and combining the inequalities yields
\[
F\!\left(Q^{(t+1)}, \tilde{Q}^{(t+1)}\right)
  \;\le\;
F\!\left(Q^{(t)}, \tilde{Q}^{(t)}\right).
\]
Since $F(Q, \tilde{Q}) \ge 0$, the sequence is bounded below and therefore convergent.

Any accumulation point must satisfy optimality with respect to both KL-projection steps. By Theorem~\ref{thm:existence-uniqueness}, the minimizer of Equation~\eqref{eq:opt_problem} is unique, and thus the entire sequence converges to $Q^\star$. The convergence statement follows directly from the Csisz\'ar-Tusn\'ady alternating minimization theorem.

\end{proof}

\section*{Appendix C \quad Algorithm}
\label{sec:appendixC}

Algorithm~\ref{alg:pid-ipfp} describes the Alternating Minimization procedure using Iterative Proportional Fitting Procedure (IPFP) in the log-domain for Sinkhorn normalization.

\begin{algorithm}[t]
\caption{Alternating Minimization for PID (IPFP-Based Solver)}
\label{alg:pid-ipfp}
\textbf{Input:} $P(x_1,x_2,y)$; $p(x_1,y)$, $p(x_2,y)$; $p(y)$ \Comment{joint empirical + marginals}

\textbf{Hyperparameters:} $T, S, \tau_{\text{outer}}, \tau_{\text{sink}}, \varepsilon>0$

\textbf{Output:} $Q^\star$, $R, U_1, U_2, S, C_1, C_2$, 

\begin{algorithmic}[1]

\State Define feasible set $\Delta_p$ (Eq.~\eqref{eq:delta_p}) and objective (Eq.~\eqref{eq:opt_problem}) \Comment{KL-projection problem}

\State \textbf{Initialization:} For each $y$
\Statex \[
Q^{(0)}(\cdot,\cdot,y)=\max\!\left(\frac{p(x_1,y)\,p(x_2,y)^\top}{p(y)},\,\varepsilon\right)
\]
\Statex \[
Q^{(0)}(\cdot,\cdot,y) \gets Q^{(0)}(\cdot,\cdot,y) \cdot \frac{p(y)}{\sum_{x_1,x_2} Q^{(0)}(x_1,x_2,y)}
\] \Comment{label-wise product initialization}

\For{$t=0$ to $T-1$}
    \State $\tilde Q^{(t)}(x_1,x_2,y)=Q^{(t)}(x_1,x_2)/|\mathcal{Y}|$ \Comment{auxiliary update (Eq.~\eqref{eq:conditional_entropy})}

    \For{each $y$ with $p(y)>\varepsilon$}
        \State $A_y\gets \tilde Q^{(t)}(\cdot,\cdot,y)$ \Comment{kernel for Sinkhorn}
        \State $\log r_y=\log(\max(p(x_1,y),\varepsilon))$; $\log c_y=\log(\max(p(x_2,y),\varepsilon))$ \Comment{stabilized marginals}

        \Repeat
            \State $\log u_y \gets \log r_y-\mathrm{LSE}_{x_2}(\log A_y+\log v_y)$
            \State $\log v_y^{\text{new}}\gets \log c_y-\mathrm{LSE}_{x_1}(\log A_y+\log u_y)$
            \State $\log v_y \gets \log v_y^{\text{new}}$
        \Until{$\|\log v_y^{\text{new}} - \log v_y\|_\infty < \tau_{\text{sink}}$} \Comment{log-domain Sinkhorn updates (Eq.~\eqref{eq:log_updates})}

        \State $Q^{(t+1)}(\cdot,\cdot,y)=\exp(\log A_y+\log u_y+\log v_y)$ \Comment{recover transport plan (Eq.~\eqref{eq:Ry_factorization})}

        \State $Q^{(t+1)}\gets\max(Q^{(t+1)},\varepsilon)$ \Comment{numerical floor}
        \State Normalize $\sum_{x_1,x_2}Q^{(t+1)}(\cdot,\cdot,y)=p(y)$ \Comment{mass-fix per label}
    \EndFor

    \State $F^{(t+1)}=\mathrm{KL}(Q^{(t+1)}\Vert \tilde Q^{(t)})$ \Comment{evaluate objective}
    \If{$\frac{|F^{(t+1)}-F^{(t)}|}{\max(1,|F^{(t)}|)}<\tau_{\text{outer}}$} \Comment{relative improvement small}
        \State \textbf{break}
    \EndIf
\EndFor

\State $Q^\star\gets Q^{(t+1)}$ \Comment{final coupling}
\State Compute $R, U_1, U_2, S$ via Eqs.~\eqref{eq:defn_r}-\eqref{eq:defn_s}\Comment{PID components}
\State $C_1 = U_1 / (U_1 + U_2)$, \quad $C_2 = U_2 / (U_1 + U_2)$ \Comment{normalized modality contributions (Eq.~\eqref{eq:Ci_normalized})}
\State \Return $Q^\star, R, U_1, U_2, S, C_1, C_2$
\end{algorithmic}
\end{algorithm}

\section*{Appendix D \quad Layerwise Analysis}
\label{sec:appendixD}

We analyzed the layerwise analysis of various models, including the SmolVLM and LLaVa, on the ScienceQA dataset and noticed some interesting details.

\subsection*{D.1 \quad LLaVa}

\begin{figure}[ht]
    \centering
    \includegraphics[width=0.48\textwidth, height=6cm, keepaspectratio]{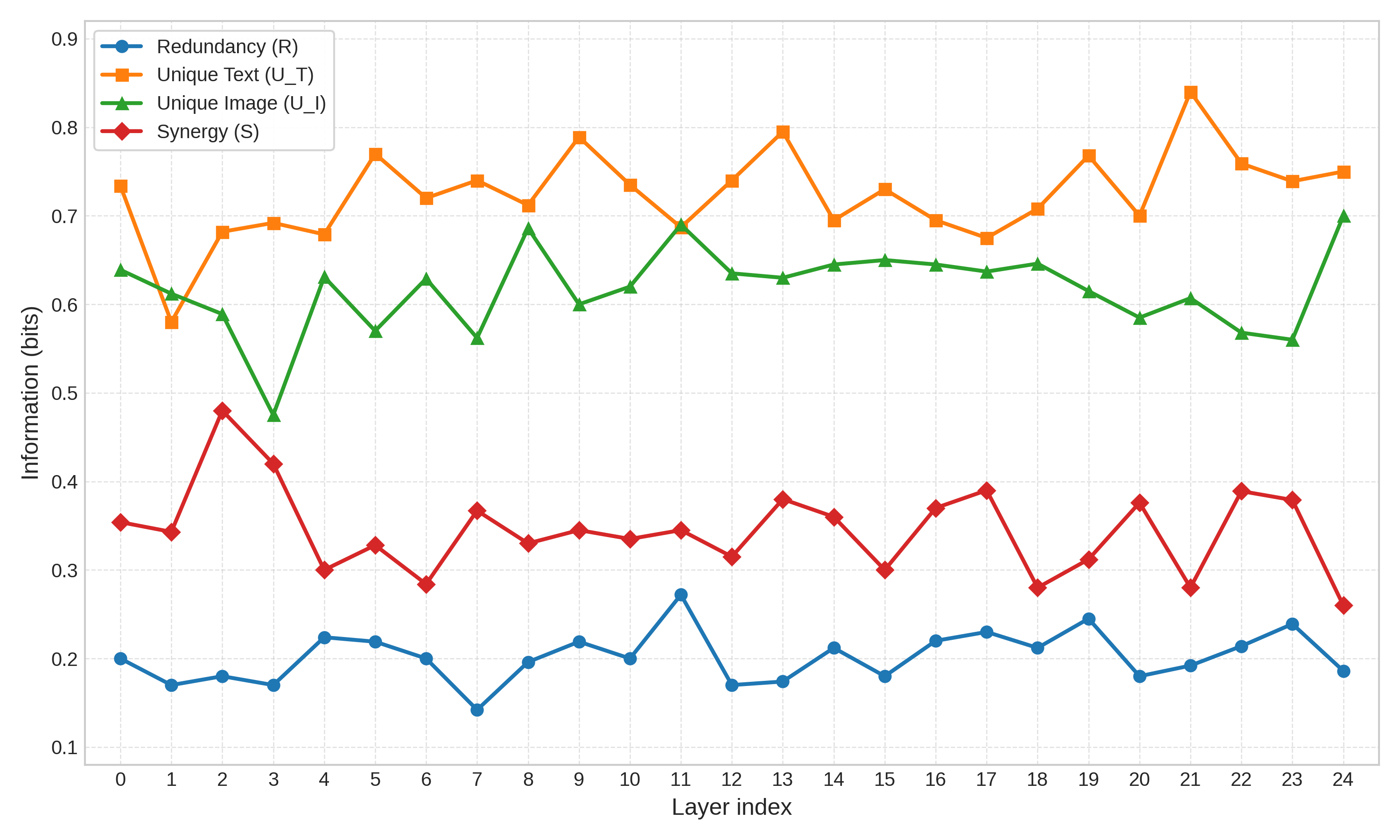}
    \hfill
    \includegraphics[width=0.48\textwidth, height=6cm, keepaspectratio]{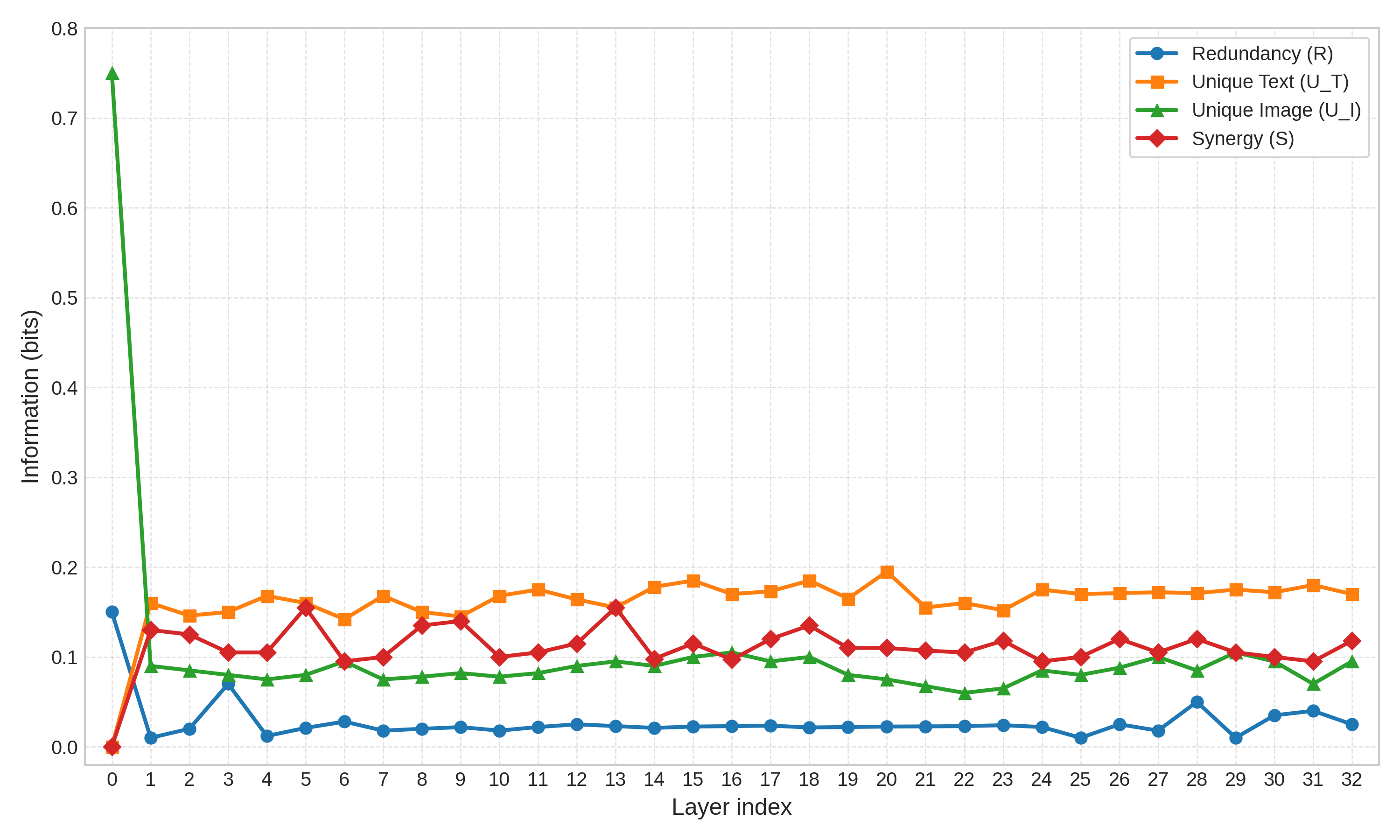}
    \vspace{0.5em} 
    \caption{Layerwise analysis of SmolVLM (left) and LLaVA (right).}
    \label{fig:layerwise-comparison}
\end{figure}

Figure \ref{fig:layerwise-comparison} (right) shows how different components of information evolve across 33 layers in the model LLaVa-7B model, highlighting the decomposition of data using the PID metrics. Redundancy (blue) peaks at the first layer and decreases quickly, suggesting early filtering of redundant information. Unique Text (orange) and Unique Image (green) remain stable across layers, with text being slightly more dominant, indicating consistent processing of both modalities. Synergy (red) fluctuates, reflecting the varying ability of the model to combine text and image information effectively at different layers. This indicates that while the model processes unique text and image data efficiently, its ability to integrate both sources fluctuates, potentially affecting performance in multimodal tasks. Overall, the model demonstrates a hierarchical approach to information processing, with challenges in maintaining balanced synergy between text and image throughout the layers.


\subsection*{D.2 \quad SmolVLM}

Figure \ref{fig:layerwise-comparison} (left) shows the layerwise PID decomposition for the SmolVLM model across 25 layers, breaking down information into redundancy, unique text, unique image, and synergy components. Redundancy (blue) remains low and stable between 0.15 and 0.25 bits, indicating consistent but moderate overlapping information without a sharp initial drop. Unique Text (orange) is the dominant component, fluctuating between 0.57 and 0.85 bits, reflecting the model’s strong and steady focus on extracting textual information. Unique Image (green) is also stable, ranging from 0.48 to 0.7 bits, showing consistent but somewhat lesser emphasis on image-specific information. Synergy (red) varies moderately between 0.25 and 0.48 bits, indicating ongoing but fluctuating integration of text and image information across layers. Overall, SmolVLM maintains steady redundancy, emphasizes unique text heavily, processes unique image data consistently, and exhibits moderate but variable multimodal fusion, suggesting a balanced and stable approach to multimodal understanding with a text-focused bias.


\section*{Appendix E \quad Complexity Analysis of IPFP}
\label{sec:appendixE}

Following the approach of ~\citet{liang2023quantifying}, the convex-programming-based estimator (CVX) computes the PID by solving a constrained convex optimization problem that maximizes the conditional entropy $\mathsf{H}_q(Y \mid X_1, X_2)$, equivalently formulated as a KL-divergence maximization over valid joint distributions $Q \in \Delta_p$ satisfying the marginal constraints. Here, $|X_1| = m$, $|X_2| = n$, and $|Y| = k$, so $Q$ is a tensor of shape $m \times n \times k$. This can be cast as a conic optimization problem and solved using interior-point-based solvers such as SCS or MOSEK. Each iteration of such solvers requires matrix factorizations involving the full tensor, leading to a worst-case complexity of $\mathcal{O}((mnk)^3)$. While this approach guarantees convergence to the global optimum, the cubic dependence on the problem size makes it impractical for larger alphabets or continuous relaxations.

In contrast, our IPFP solver adopts an alternating Sinkhorn-style update scheme that enforces the marginal constraints through multiplicative scaling, avoiding expensive global factorizations. Each outer iteration updates all $k$ conditional slices and performs approximately $S$ inner Sinkhorn updates per slice, resulting in a per-iteration cost of $\mathcal{O}(Smnk)$ and memory complexity of $\mathcal{O}(mnk)$. In practice, the IPFP procedure converges within a few dozen iterations, achieving a near-identical solution to the CVX estimator while reducing computational cost by several orders of magnitude.

\section*{Appendix F \quad Visualization}
\label{sec:appendixF}

To qualitatively examine how the model distributes information across modalities, Table~\ref{case_study} visualizes several examples produced by the Paligemma-3B model on the GQA dataset. Each sample includes the input image, the corresponding question and model response, the PID metrics: redundant (\textbf{R}), unique information from text (\textbf{U1}), unique information from image (\textbf{U2}), and synergistic information (\textbf{S}), as well as the model confidence scores for the text modality (\textbf{C1}) and the image modality (\textbf{C2}).

\begin{table*}[htp]
\centering
\fontsize{9pt}{11pt}\selectfont
\renewcommand{\arraystretch}{1.3}
\begin{tabular}{@{} >{\centering\arraybackslash}m{0.3\linewidth} 
                    >{\centering\arraybackslash}m{0.3\linewidth} 
                    >{\centering\arraybackslash}m{0.4\linewidth} @{}}
\toprule
\textbf{Image} & \textbf{Question and Model Response} & \textbf{PID Metrics and Model Confidence Scores} \\
\midrule

{\centering
\includegraphics[width=0.3\textwidth]{figures/viz_image_0.png}
\par}
&
{\centering
\begin{tabular}{@{}p{\linewidth}@{}}
\textbf{User:} Is it overcast? \\
\textbf{PaliGemma-3B:} No
\end{tabular}
\par}
&
{\centering
\setlength{\tabcolsep}{3pt} 
\begin{tabular*}{\linewidth}{@{\extracolsep{\fill}} cccccc @{}}
\toprule
\textbf{R} & \textbf{U1} & \textbf{U2} & \textbf{S} & \textbf{C1} & \textbf{C2} \\
\midrule
0.3466 & 0.3465 & 0.857 & 0.001 & 0.567 & 0.480 \\
\bottomrule
\end{tabular*}
\par}
\\[2ex] 

{\centering
\includegraphics[width=0.3\textwidth]{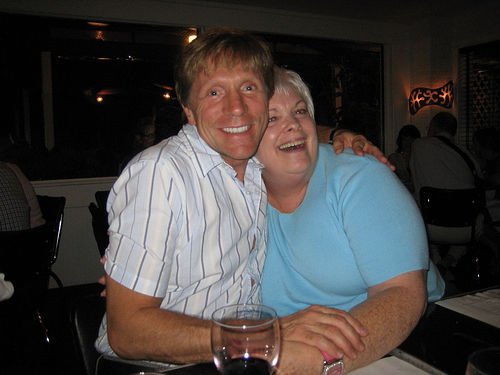}
\par}
&
{\centering
\begin{tabular}{@{}p{\linewidth}@{}}
\textbf{User:} Who is wearing the dress? \\
\textbf{PaliGemma-3B:} Woman
\end{tabular}
\par} 
&
{\centering
\setlength{\tabcolsep}{3pt} 
\begin{tabular*}{\linewidth}{@{\extracolsep{\fill}} cccccc @{}}
\toprule
\textbf{R} & \textbf{U1} & \textbf{U2} & \textbf{S} & \textbf{C1} & \textbf{C2} \\
\midrule
0.3635 & 0.2772 & 0.857 & 0.0001 & 0.5673 & 0.43226 \\
\bottomrule
\end{tabular*}
\par} 
\\[2ex]

{\centering
\includegraphics[width=0.3\textwidth]{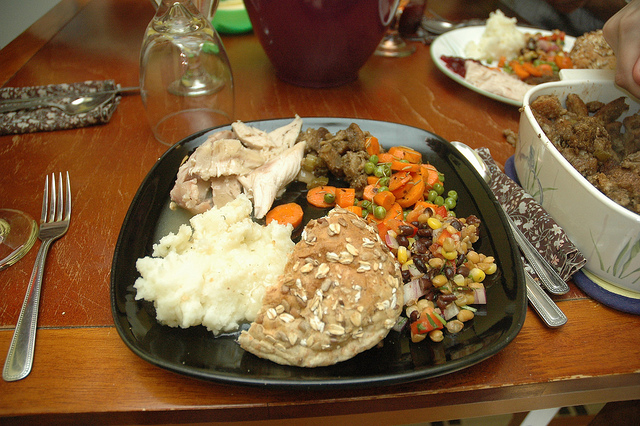}
\par}
&
{\centering
\begin{tabular}{@{}p{\linewidth}@{}}
\textbf{User:} Does the utensil on top of the table look clean and black? \\
\textbf{PaliGemma-3B:} No
\end{tabular}
\par}
&
{\centering
\setlength{\tabcolsep}{3pt} 
\begin{tabular*}{\linewidth}{@{\extracolsep{\fill}} cccccc @{}}
\toprule
\textbf{R} & \textbf{U1} & \textbf{U2} & \textbf{S} & \textbf{C1} & \textbf{C2} \\
\midrule
 0.432 & 0.399 & 0.9464 & 0.1249 & 0.5198 & 0.480 \\
\bottomrule
\end{tabular*}
\par} 
\\[2ex]

{\centering
\includegraphics[width=0.3\textwidth]{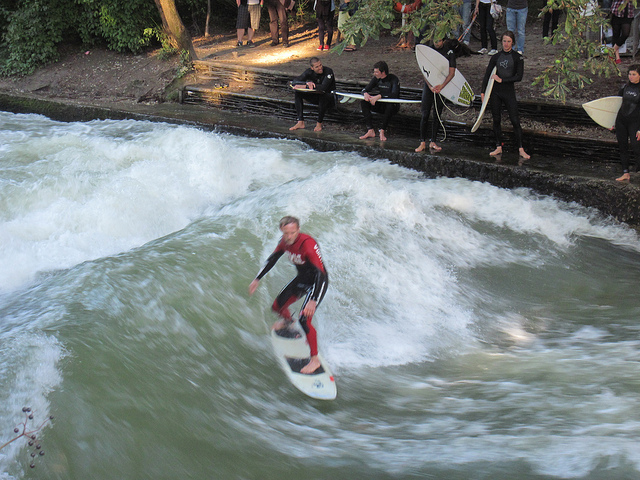}
\par}
&
{\centering
\begin{tabular}{@{}p{\linewidth}@{}}
\textbf{User:} Is the surfer that looks wet wearing a wetsuit? \\
\textbf{PaliGemma-3B:} Yes
\end{tabular}
\par}
&
{\centering
\setlength{\tabcolsep}{3pt} 
\begin{tabular*}{\linewidth}{@{\extracolsep{\fill}} cccccc @{}}
\toprule
\textbf{R} & \textbf{U1} & \textbf{U2} & \textbf{S} & \textbf{C1} & \textbf{C2} \\
\midrule
0.3696 & 0.6308 & 0.5241 & 0.1848 & 0.3694 & 0.6305 \\
\bottomrule
\end{tabular*}
\par} 
\\[2ex]

{\centering
\includegraphics[width=0.3\textwidth]{figures/viz_image_4.png}
\par}
&
{\centering
\begin{tabular}{@{}p{\linewidth}@{}}
\textbf{User:} Are there clouds in the sky? \\
\textbf{PaliGemma-3B:} Yes, slightly.
\end{tabular}
\par}
&
{\centering
\setlength{\tabcolsep}{3pt} 
\begin{tabular*}{\linewidth}{@{\extracolsep{\fill}} cccccc @{}}
\toprule
\textbf{R} & \textbf{U1} & \textbf{U2} & \textbf{S} & \textbf{C1} & \textbf{C2} \\
\midrule
0.4123 & 0.3521 & 0.841 & 0.002 & 0.592 & 0.468 \\
\bottomrule
\end{tabular*}
\par} 
\\[2ex]

\end{tabular}
\caption{\label{case_study}
Examples of PID metrics and model confidence scores produced by the PaliGemma-3B model on selected samples from the GQA dataset.
}
\end{table*}

\end{document}